\documentclass[letterpaper]{article}
\usepackage{aaai}
\usepackage{times}
\usepackage{helvet}
\usepackage{courier}
\usepackage{color}
\usepackage{comment}
\usepackage{multirow}
\usepackage{amsmath}
\usepackage{amsfonts}
\usepackage{graphicx}
\usepackage{bm}
\usepackage{CJKutf8}
\begin{document}
%
\title{Character-level Japanese Text Generation with\\ Attention Mechanism for Chest Radiography Diagnosis}
\author{Kenya Sakka\textsuperscript{\rm 1}, Kotaro Nakayama\textsuperscript{\rm 1,2}, Nisei Kimura\textsuperscript{\rm 1}, Taiki Inoue\textsuperscript{\rm 1}, Yusuke Iwasawa\textsuperscript{\rm 1},
\\ \Large \textbf{Ryohei Yamaguchi\textsuperscript{\rm 1}, Yoshimasa Kawazoe\textsuperscript{\rm 1}, Kazuhiko Ohe\textsuperscript{\rm 1}, Yutaka Matsuo\textsuperscript{\rm 1}}\\ 
\textsuperscript{\rm 1}The University of Tokyo \textsuperscript{\rm 2}NABLAS Inc. \\ 
7-3-1 Hongo, Bunkyo-ku, Tokyo, 113-8586, Japan\\
sakka\_kenya\_17@stu-cbms.k.u-tokyo.ac.jp, \{nakayama, kimura, iwasawa, matsuo\}@weblab.t.u-tokyo.ac.jp\\
taiki-inoue@g.ecc.u-tokyo.ac.jp, r-yamagu@m.u-tokyo.ac.jp, \{kawazoe, kohe\}@hcc.h.u-tokyo.ac.jp
}
\maketitle
\begin{abstract}
Chest radiography is a general method for diagnosing a patient's condition and identifying important information; therefore, radiography is used extensively in routine medical practice in various situations, such as emergency medical care and medical checkup. However, a high level of expertise is required to interpret chest radiographs. Thus, medical specialists spend considerable time in diagnosing such huge numbers of radiographs. In order to solve these problems, methods for generating findings have been proposed. However, the study of generating chest radiograph findings has primarily focused on the English language, and to the best of our knowledge, no studies have studied Japanese data on this subject. There are two challenges involved in generating findings in the Japanese language. The first challenge is that word splitting is difficult because the boundaries of Japanese word are not clear. The second challenge is that there are numerous orthographic variants. For deal with these two challenges, we proposed an end-to-end model that generates Japanese findings at the character-level from chest radiographs. In addition, we introduced the attention mechanism to improve not only the accuracy, but also the interpretation ability of the results. We evaluated the proposed method using a public dataset with Japanese findings. The effectiveness of the proposed method was confirmed using the Bilingual Evaluation Understudy score. And, we were confirmed from the generated findings that the proposed method was able to consider the orthographic variants. Furthermore, we confirmed via visual inspection that the attention mechanism captures the features and positional information of radiographs.
\end{abstract}

\section{Introduction}
Medical imaging is widely used for the diagnosis and treatment of several diseases. In particular, chest radiography is a general method for determining the patient's condition and for identifying important information. Thus, chest radiography is performed for a large number of patients in several situations. such as emergency medical care and medical checkup. However, a high level of expertise is required for interpreting chest radiographs. Thus, a large workforce of medical specialists is necessary. As a result, medical workers including radiologists are over-burdened with work, and solutions are required to resolve this issue. 

\begin{figure}[t!]
\centering
\includegraphics[scale=0.2]{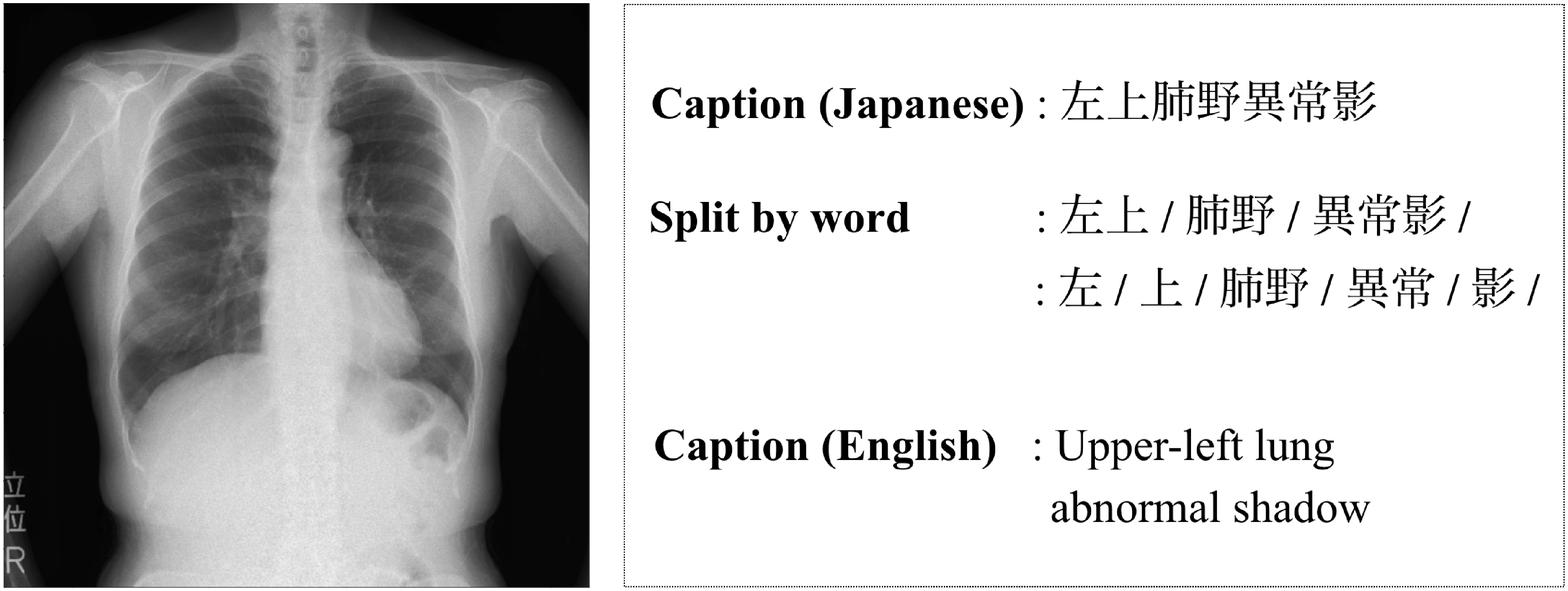}
\caption{An example of medical image captions in Japanese. The top caption is the Japanese caption for the image. The middle caption presents some examples of Japanese captions split into words. The bottom caption is the English notation corresponding with the top caption.}
\label{tab:dlb_contents}
\end{figure}

Therefore, research has been actively conducted to automatically detect findings as a classification problem using image recognition technology \cite{Wang-1, Rajpurkar, Baltruschat, Wong}. However, the mechanism of the human body is complicated; also, classification tasks that categorize patients into predetermined classes are challenging. Therefore, in the previous one or two years, an end-to-end model has also been proposed that outputs text information of the findings directly, based on the images \cite{Jing, Wang-2, Li-1, Christy-2}. However, the study of generating chest radiograph findings primarily focused on the English language.

The Japanese Social of Radiological Technology (JSRT) Dataset \cite{Junji} created by JSRT consists of Japanese chest radiographs (Figure 1). Research on the classification of chest radiographs using the JSRT Dataset has been conducted \cite{Hamada}. However, to the best of our knowledge, no study has been conducted to target findings generation in the Japanese language. Moreover, the linguistic features of the text also vary for each region. Thus, it is challenging to use the existing model for establishing a diagnosis on the basis of the chest radiographs in the Japanese language. Furthermore, there are two challenges involved in generating findings in the Japanese language. The first challenge is that word splitting is difficult (Figure 1, middle caption). For example, ``Upper-left lung abnormal shadow" is expressed as \begin{CJK}{UTF8}{ipxm}``左上肺野異常影"\end{CJK} in the Japanese language. A sentence in the English language can be split into words based on the spaces. However, the boundaries of words are unclear in the Japanese language. In general, word splitting in the Japanese language is conducted by matching words in sentences using a vocabulary dictionary \cite{Kudo}. The findings of medical images mainly consist of technical terms that are not supported by the existing vocabulary dictionaries. The second challenge is that there are many orthographic variants of words in the Japanese language. In the case of findings on chest radiographs, when the abnormal area is on both sides of the lungs, there are multiple expressions, such as \begin{CJK}{UTF8}{ipxm}``両肺", ``両側肺", ``両肺部", and ``両側肺部"\end{CJK} (English: both sides of the lung). For overcoming these challenges, complicated pre-processing  (e.g., preparation of a vocabulary dictionary and morphological analysis) is required, and the versatility of the method is low. 

Here, we proposed an Encoder-Decoder model with attention mechanism that inputs chest radiographs and generates findings in the Japanese language. In generating the findings in the Japanese language, we challenged the character-level \cite{Bandanau, Mao} approach to deal with the complexity of word splitting and orthographic variants of words. The character-level approach has a feature that does not need to create a dictionary of special words. Therefore, the character-level approach is a robust against changes in finding data and notation.

We evaluated the proposed method using a public JSRT Dataset. The effectiveness of the proposed method was confirmed using the BLEU score \cite{Kishore}. We found that the proposed method could generate Japanese findings corresponding to the orthographic variants. Furthermore, we have confirmed via visual inspection that the attention mechanism captures the features and positional information of chest radiographs.

Overall, the main contributions of our work are as follows:
\begin{itemize}
\item We proposed a method for generating character-level Japanese findings from chest radiographs.
\item We generated character-level findings to overcome the challenges experienced in splitting Japanese language words and the orthographic variants of words.
\item We introduce the attention mechanism at the character level for improving the accuracy and interpretation of the results.
\end{itemize}

\begin{figure*}[t]
\begin{center}
\includegraphics[scale=0.38]{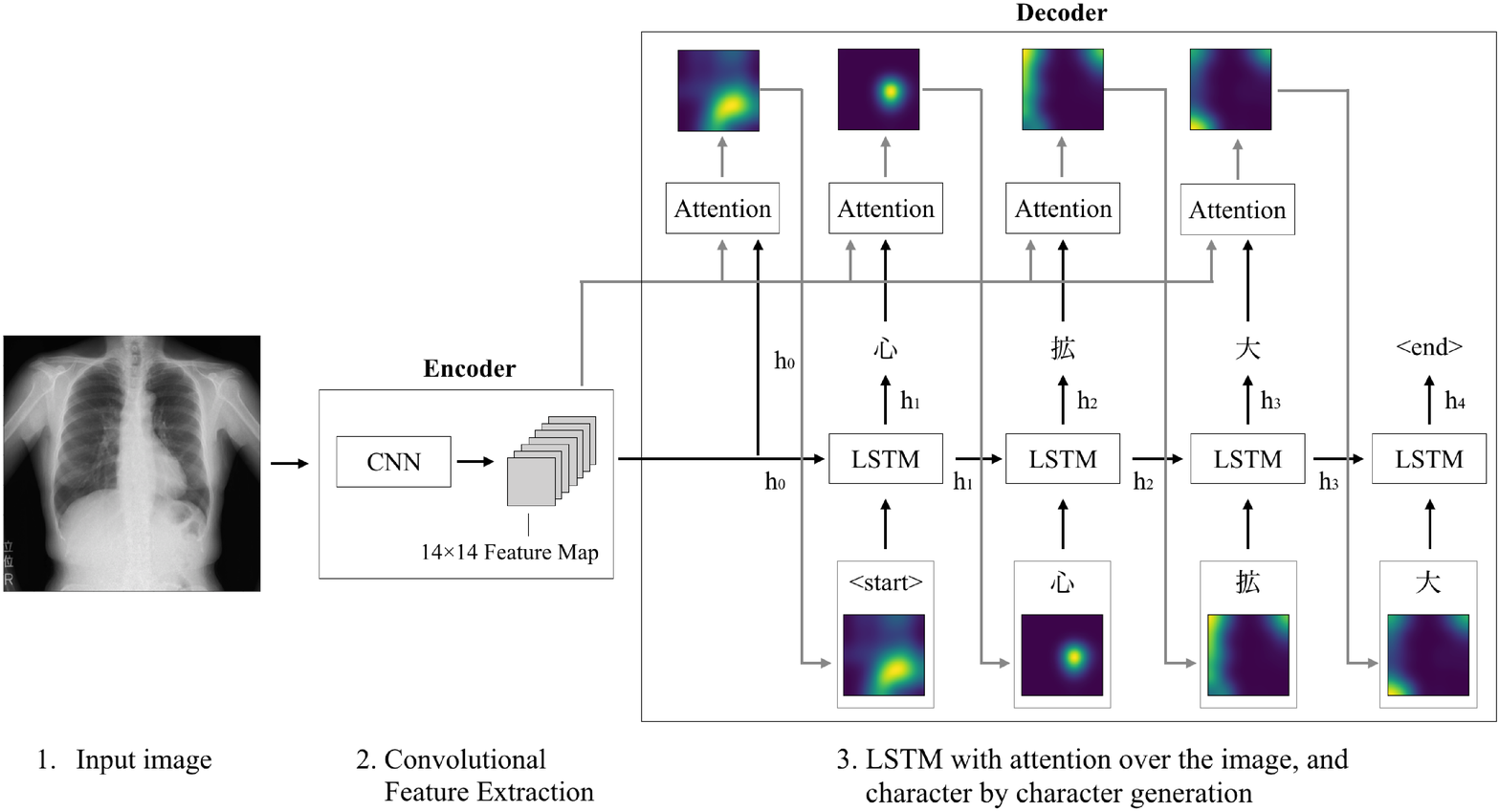}
\end{center}
\caption{Architecture of the proposed model. (1) Input the pre-processed chest radiograph into the encoder (details are given in the Encoder section). (2) The feature maps for the input image are extracted using Convolutional Neural Network (CNN). The size of each feature map is $14 \times 14$ and the number of channels is 2048. (3) In the decoder, the findings are generated at the character-level using long short-term memory (LSTM) and the attention mechanism based on the extracted feature maps. When the $\langle end \rangle$ symbol is output or the character length exceeds a certain length, the generation of findings is stopped.}
\end{figure*}


\section{Related Work} 

\textbf{Medical image analysis with machine learning.} \quad
Machine learning for health care has been extensively researched in the fields of academics and industries. In particular, in the area of medical images, applications are being made to reduce the burden on medical professionals. The CheXNet \cite{Rajpurkar} resolved the classification problems for chest radiographs. The classification accuracy in 14 classes exceeded that of medical professionals. As a further application, the automatic generation of medical image reports or findings is gaining increasing research interest. However, previous studies have focused on English sentences and generated medical reports at the word or sentence level \cite{Xue, Li-1}. In particular, the generation of medical reports using attention mechanism \cite{Jing}, tags for findings are used to improve accuracy. The tags are not attached in a normal diagnosis; therefore, they need to be given manually by a medical professional.
In previous studies, the target language of the generated medical report is limited to the English language, and to the best of our knowledge, no study has been conducted with the objective of generating Japanese findings. Chest radiographs obtained in Japan are captioned in the Japanese language. Therefore, it is necessary to construct a model that generates Japanese medical findings suitable for the Japanese language structure.\\

\noindent \textbf{Character-level language model.} \quad
In the field of natural language processing and caption generation, a language model has generally used that learned distributed representation at the word-level \cite{Bengio, Mikolov}. However, word-level language models have problems, such as the need for pre-processing (e.g., word splitting and preparation of a vocabulary dictionary) and difficulty to cope with orthographic variants. There are many abbreviated expressions in the Japanese language, for example, \begin{CJK}{UTF8}{ipxm}``両側肺部"\end{CJK} and \begin{CJK}{UTF8}{ipxm}``両肺"\end{CJK} (English: Both sides of the lung). Therefore, a character-level language model has been proposed in recent years \cite{Kim, Saxe, Rami}. The character-level language model is also effective when there are orthographic variants such as user-generated content \cite{Zhang}. In addition, a character-level language model can be used without preparing an additional vocabulary dictionary even in highly specialized fields because there is no requirement for word splitting. In the Japanese language, a character-level language model is effective because word splitting is more difficult than in the English language (Figure 1). A character-level language model was difficult to apply in the general captioning problem, because the model required to learn long-term relationships among the characters. However, the findings of the medical images are shorter than the general caption datasets such as MS COCO \cite{Lin}. In addition, Japanese text contains more position and context information than other languages. For example, the word \begin{CJK}{UTF8}{ipxm}``右"\end{CJK} in the Japanese language represents the word ``right" in the English language. In this case, one Japanese character has as much information as five English characters. Therefore, a character-level language model is an effective method for generating the medical findings in the Japanese language.\\

\noindent \textbf{Image captioning.} \quad
Image captioning aims to automatically generate natural language description from images. The main objective of image captioning is to use CNN for image classification and recurrent networks for sequence modeling \cite{Mao, Vinyals, Fang, Donahue}.

Recently, the method of image captioning using attention mechanisms is shown to be useful \cite{Kelvin, You}. Attention mechanisms not only improve the accuracy, but also increase the ease of interpretation of the results by visualizing ``where'' and ``what'' the attention is focused. Attention mechanisms create weights based on the feature maps extracted using CNN. Particularly, soft-attention is calculated attention weight by weighing and adding together each channel of feature maps. Therefore, the attention mechanism functions in a wide range. To interpret a chest radiograph, it is necessary to refer to various regions in the image. Therefore, here, we generated medical findings using soft-attention, that allows the attention mechanism to function in a wide range.

\section{Methods}

\textbf{Overview.} \quad 
In this paper, we propose an Encoder-Decoder model that uses a soft-attention mechanism to output the medical findings of the Japanese language at character-level. A chest radiograph is input into the encoder model. Thereafter, feature maps are extracted using CNN. The decoder model generates medical findings at character-level by using the feature map extracted from the encoder model.\\

\noindent \textbf{Pre-training.} \quad
We extracted feature maps using ResNet-151 \cite{Kaiming}. The model of ResNet-151 is pre-trained with ImageNet \cite{Deng}. ImageNet is a dataset for general object images; thus, a pre-trained model using ImageNet does not always extract appropriate feature maps from medical images. Therefore, it is necessary to pre-train ResNet-151 using chest radiographs. In this paper, we prepared 13,144 chest radiographs as the dataset of a pre-trained model from the JSRT dataset. The dataset comes with one of 11 species label per image and split so that 80\% as training data and 20\% as validation data. The encoder was trained by solving a classification problem of all 11 classes. The data of pre-training is not included in the test dataset for evaluating a character-level finding generation model.\\

\noindent \textbf{Encoder.} \quad
Our encoder model takes three channel chest radiographs. Each channel consists of the same gray scale chest radiograph. We extracted feature maps using pre-trained ResNet-151. The extractor produces $L$ vectors, each of which is a D-dimensional representation corresponding to a part of the image. In this paper, D is $14 \times 14$ and $L$ is 2048.
\begin{equation}
a = \{\bm{a_{1}}, \dots, \bm{a_{L}} \}, \bm{a_{i}} \in \mathbb{R}^{D}
\end{equation}

The following pre-processing was conducted when the chest radiograph was input into the encoder model. The size of the chest radiograph was too large ($2048 \times 2048 \times 3$); therefore, the encoder could not be trained, and the appropriate features could not be extracted. Therefore, the chest radiograph was resized to $256 \times 256 \times 3$, and Random Crop was conducted to make the size $224 \times 224 \times 3$. Thereafter, the values of each channel in the image were normalized to the mean (0.485, 0.456, 0.406) and standard deviation (0.229, 0.224, 0.225) \cite{Deng, Rajpurkar}.\\

\noindent \textbf{Decoder.} \quad 
The decoder consists of a single layer LSTM \cite{Hochreiter}, and a medical finding $y$ was generated by outputting one character at each step.
\begin{equation}
y = \{\bm{y_{1}}, \dots, \bm{y_{C}} \}, \bm{y_{i}} \in \mathbb{R}^{K}
\end{equation}
where, $y_{i}$ is each generated character, $K$ is vocabulary size, and $C$ is the length of medical findings.

LSTM takes three inputs: the context vector $z_{t}$, the hidden state of the previous step $h_{t-1}$, and the character generated in the previous step $y_{t-1}$. The initial value of $h_{0}$ is zero vector. The context vector $z$ capturing the visual information associated with a particular input location, then $z_{t}$ is a dynamic representation of the relevant part of the image input at time $t$. We define a mechanism $\Phi$ that computes $z_{t}$ from annotation vectors $a_{i}, i=1,\ldots, L$ corresponding to the features extracted at different image locations \cite{Bandanau, Kelvin} We generated $\alpha_{i}$ with a positive weight for each $i$ that functions to focus correctly on the position in the image and support the generation of the next character. The weight $\alpha_{i}$ of each annotation vector $a_{i}$ was calculated by the attention model $f_{att}$ subject to the hidden state of the previous step $h_{t-1}$.
\begin{eqnarray}
e_{ti} &=& f_{att}(\bm{a_{i}}, \bm{h_{t-1}})\nonumber\\
&=&\bm{W}_{t_{att}}ReLU(\bm{W_{t}a_{i}}+\bm{W_{t, h}h_{t-1}}+\bm{b_{t}})\\
\nonumber\\
\alpha_{ti} &=& softmax_{t}(e_{ti}) = \frac{\exp{e_{ti}}}{\sum_{k=1}^{L}\exp(e_{tk})}
\end{eqnarray}

where $\bm{W}_{t_{att}}$, $\bm{W}_{t}$ and $\bm{W}_{t, h}$ denotes parameter matrices and $\bm{b_{t}}$ denotes bias of the attention model.

when the weight $\alpha$ is calculated, the context vector $z_{t}$ is calculated as follows.
\begin{eqnarray}
\bm{z_{t}} &=& \Phi(\bm{a_{i}}, \alpha_{i})\\
\Phi(a_{i}, \alpha_{i}) &=& \sum_{i=1}^{L}\bm{a_{i}}\alpha_{i}
\end{eqnarray}

$\Phi$ is a function that takes a pair of weights $\alpha_{i}$ corresponding to the annotation vector $a_{i}$ and returns a one-dimensional vector.

The model generates the character $y_{t}$ for each step as follows. First, input $z_ {t}$, $y_ {t-1}$ and $h_ {t-1}$ into LSTM to calculate the hidden state $h_ {t}$ of the current step. The calculated $h_{t}$ was input to one fully connected layer with a dropout layer, and the character with the highest generation probability was used as the output $y_{t}$ of the current step.

The hyper parameters of decoder were set as follows. The hidden state of LSTM was 256 dimensions, the dropout rate was 0.5 \cite{Nitish}, and the optimization function was Adam \cite{Diederik}. In this paper, we did not use distributed representation because of the low number of characters.

The proposed method can generate the medical findings that consider the previous state in more detail by using an attention mechanism for each character. In addition, the proposed method improves the interpretability of the results by introducing an attention mechanism.\\

\noindent \textbf{Learning condition.} \quad
The proposed model is trained end-to-end by minimizing the following loss function.
\begin{equation}
Loss = -\sum_{k}^{K}t_{k}\log y_{k} + \lambda \sum_{i}^{L} (1-\sum_{t}^{C}\alpha_{ti})^{2}
\end{equation}

The first term of the loss function represents the cross-entropy error of the characters generated at each step. The second term is a regularization term to make the attention mechanism work equally on each channel of the feature map. $\lambda$ adjusts the priority of the second term, and we set $\lambda$ to 1.0 \cite{Kelvin}. The hyper parameters were set as follows, and the model training was conducted. Batch size was 16, dropout rate of the decoder is 0.5, learning rate of the encoder is 0.0001, and the learning rate of the decoder was 0.0004. When the validation data was not improved for 10 epochs continuously, the learning rate was adjusted by multiplying the learning rate of the encoder and the decoder by 0.8 \cite{Anders}. The maximum number of epoch was 200, and early stopping \cite{Rich} was conducted when the BLEU-4 score showed no improvement for 20 consecutive epochs of the validation data to avoid overfitting.\\

\noindent \textbf{Prediction.} \quad
Prediction for new data was performed as follows using the trained encoder and decoder. First, we resized an input image of $2048 \times 2048 \times 3$ to $224 \times 224 \times 3$ and normalized in the same method as training. We obtained feature maps of size $14 \times 14 \times 2048$ by inputting in the pre-processing image into the encoder. The average value of the obtained feature maps for each channel was used as the initial value of the context vector $z$, and the medical finding was generated by inputting context vector into the decoder together with the $\langle start \rangle$ symbol indicating the start of generation. Each step of the decoder takes as input, the character generated one step before $y_{t-1}$ and the context vector of the current step $z_{t}$. The context vector of the current step $z_{t}$ is calculated by feature maps $a$ and the hidden state of one step before $h_{t-1}$ (Eq. 5). And the encoder generates a character of the current step $y_{t}$. The prediction ends when the $\langle end \rangle$ symbol indicating the end of medical finding generation or the character length of the generated medical finding exceeds a certain length. In the generation of medical findings of Beam Size set for each step was generated by using Beam Search \cite{Waiseman}, and the prediction result was the one with the highest probability of finding at the end of prediction step.


\section{Experiments}
\noindent \textbf{Datasets.} \quad
We evaluated the proposed method using the JSRT Dataset. The JSRT Dataset is a public dataset that contains 18,004 pairs of frontal chest radiographs and Japanese findings. We prepared two datasets based on the threshold for the appearance frequency of the medical findings (Table 1). The data with the appearance frequency of finding lower than the threshold were excluded because the model cannot learn the features of data sufficiently \cite{Jing, Christy-2}. The thresholds were set at 5 and 30. A dataset with a threshold of 5 covers 94.79\% of all data and includes 164 findings composed of 118 characters. The minimum number of characters in the findings was 3 and the maximum number of characters was 9. Futhermore, a dataset with a threshold of 30 covers 87.02\% of all the data, and includes 26 findings composed of 65 characters. The minimum number of characters in the findings was 3, and the maximum number of characters was 13.

We excluded those data from our dataset that require time-series information for interpretation, such as \begin{CJK}{UTF8}{ipxm}``前回と変化なし\end{CJK} (English: No change from the previous time)" because time-series information cannot be detected using the proposed method. In addition, we formatted the dataset except for unnecessary data, such as \begin{CJK}{UTF8}{ipxm}``手入力 部位\end{CJK} (English: manual input, part)". When multiple findings were associated with a single image, the findings were assigned to the same image one by one to create multiple data.\\

\begin{table}[t]
\caption{Details regarding our datasets}
\begin{center}
\scalebox{0.8}
{
\begin{tabular}{c c c|c c c}
\hline
\multirow{2}{*}{Threshold} & \multirow{2}{*}{\begin{tabular}[c]{@{}c@{}}Number of \\ findings\end{tabular}} & \multirow{2}{*}{\begin{tabular}[c]{@{}c@{}}Number of\\ character\end{tabular}} & \multicolumn{3}{c}{Number of records}\\ \cline{4-6} &  &  & Train & \multicolumn{1}{l}{Validation} & \multicolumn{1}{l}{Test} \\ \hline
\multicolumn{1}{c}{5}    & 164 & 118 & \multicolumn{1}{c}{12,810} & 1,602 & 1,602 \\ 
\multicolumn{1}{c}{30}   & 26  & 65  & \multicolumn{1}{c}{11,752} & 1,453 & 1,453 \\ \hline
\end{tabular}
}
\end{center}
\end{table}

\begin{table*}[t]
\caption{BLEU score of the proposed methods. We evaluated the Original Distribution Dataset and Only Abnormal Dataset in oversampling and undersampling. Each dataset was filtered as per the threshold of the appearance frequency of the findings. BLEU-n indicates the BLEU score of the test data that uses up to n-grams. The last column indicates the number of findings generated with each method.}
\begin{center}
\small
{
\begin{tabular}{cccccccc}
\hline
Datasets & Methods & Threshold & BLEU-1 & BLEU-2 & BLEU-3 & BLEU-4  & Number of findings\\ \hline
\multicolumn{1}{c|}{\multirow{4}{*}{\begin{tabular}[c]{@{}c@{}}Original Distribution Dataset\end{tabular}}} & \multicolumn{1}{c|}{\multirow{2}{*}{Oversampling}} & \multicolumn{1}{c|}{5} & \textbf{0.7708} & \textbf{0.7668} & \textbf{0.7604} & \textbf{0.7535} & \multicolumn{1}{|c}{72}\\
\multicolumn{1}{c|}{} & \multicolumn{1}{c|}{} & \multicolumn{1}{c|}{30} & \textbf{0.8322} & \textbf{0.8304} & \textbf{0.8261} & \textbf{0.8214} & \multicolumn{1}{|c}{22}\\ \cline{2-8} 
\multicolumn{1}{c|}{} & \multicolumn{1}{c|}{\multirow{2}{*}{Undersampling}} & \multicolumn{1}{c|}{5} & 0.5472 & 0.5353 & 0.5091 & 0.4934 & \multicolumn{1}{|c}{42}\\
\multicolumn{1}{c|}{} & \multicolumn{1}{c|}{} & \multicolumn{1}{c|}{30} & 0.6203 & 0.6142 & 0.5962 & 0.5873 & \multicolumn{1}{|c}{26}\\ \hline \hline
\multicolumn{1}{c|}{\multirow{4}{*}{\begin{tabular}[c]{@{}c@{}}Only Abnormal Dataset\end{tabular}}} & \multicolumn{1}{c|}{\multirow{2}{*}{Oversampling}} & \multicolumn{1}{c|}{5} & 0.2348 & 0.2193 & 0.1980 & \textbf{0.1691} & \multicolumn{1}{|c}{61} \\
\multicolumn{1}{c|}{} & \multicolumn{1}{c|}{} & \multicolumn{1}{c|}{30} & 0.1980 & 0.1882 & 0.1664 & 0.1376 & \multicolumn{1}{|c}{16}\\ \cline{2-8} 
\multicolumn{1}{c|}{} & \multicolumn{1}{c|}{\multirow{2}{*}{Undersampling}} & \multicolumn{1}{c|}{5} & \textbf{0.2612} & \textbf{0.2314} & \textbf{0.2093} & 0.1439 & \multicolumn{1}{|c}{33}\\
\multicolumn{1}{c|}{} & \multicolumn{1}{c|}{} & \multicolumn{1}{c|}{30} & \textbf{0.2641} & \textbf{0.2468} & \textbf{0.2258} & \textbf{0.1713} & \multicolumn{1}{|c}{23}\\ \hline
\end{tabular}
}
\end{center}
\end{table*}

\noindent \textbf{Training detail.} \quad 
The model was trained on the imbalanced data using oversampling and undersampling \cite{Haibo, Kevin}. The data of abnormal findings indicates that is annotated with findings other than \begin{CJK}{UTF8}{ipxm}``異常なし (English: normal)"\end{CJK}. (i) Oversampling: Oversampling was performed so that the number of data for each abnormal finding was 100. Thereafter, random sampling was performed from the normal data for each epoch so that the total number of abnormal data was equal to the total number of normal data. The total number of training data was 33,206 when threshold was 5 and 5,606 when the threshold was 30. (ii) Undersampling: Random sampling was conducted for each epoch so that the normal data was equal to the average value of the number of each abnormal data. The average value for each dataset was 19 when the threshold was 5 and 80 when the threshold was 30. The total number of training data was 3,089 when the threshold was 5 and 2,092 when the threshold was 30.\\

\noindent \textbf{Evaluation metrics.} \quad
The proposed model was trained using oversampling and undersampling, and evaluated using the BLEU 1-4 score. Two types of datasets were prepared as test datasets. The first test dataset was the original distribution dataset. In the original distribution dataset, the distribution of each finding is actually obtained in the medical field. Most findings obtained in the actual medical field are normal. Therefore, ``\begin{CJK}{UTF8}{ipxm}異常なし\end{CJK} (English: normal)" accounted for approximately 75.58\% of the total data in the original distribution dataset. The second test dataset contained only abnormal data. The only abnormal dataset is prepared by excluding the data annotated by ``\begin{CJK}{UTF8}{ipxm}異常なし\end{CJK} (English: normal)". By evaluating with the only abnormal dataset, it is possible to prevent a model biased toward ``\begin{CJK}{UTF8}{ipxm}異常なし\end{CJK} (English: normal)" from being highly evaluated.


\section{Results}
The result of the attention mechanism for each character and the generated findings are shown in Figure 3. The upper-left image of each result represents the input image, and the lower-left image represents the summation of the weights of the attention mechanism generated at each step. In other words, the lower-left image represents the weight distribution of the attention mechanism at all the steps of finding generation. The right side of each result shows the weight of the attention mechanism corresponding to the generated character. Each of the below results is the answer of the finding in Japanese and English languages. In addition, the finding generated using the proposed method is also shown. A chest radiograph of the patient with the implanted device confirms that the weight distribution of the attention mechanism is high around the device when the findings were generated (Figure 3A). The attention mechanism works strongly at the position of the device throughout the generation of findings. In case where the positional information was included in the findings, such as ``\begin{CJK}{UTF8}{ipxm}両側肺尖胸肥厚\end{CJK} (English: bilateral pulmonary apical chest thickening)" and ``\begin{CJK}{UTF8}{ipxm}左助骨横隔膜角鈍\end{CJK} (English: Costophrenic angle dull)" (Figure 3B, C). We can conclude that the attention mechanism functions to reflect the positional information. Some of the findings generated by the model were confirmed to give additional information to the answer findings. For example, the proposed method generated the finding of ``\begin{CJK}{UTF8}{ipxm}軽度心拡大\end{CJK} (English: slight heart enlargement)" for the data with the answer label ``\begin{CJK}{UTF8}{ipxm}心拡大\end{CJK} (English: heart enlargement)" (Figure 3D). An example of the case where the proposed method generates incorrect findings is shown (Figure 3E). The findings generated in this example narrow down the position information to the lower-left. The answer label is the left diaphragm elevation, so the abnormal region is located in the lower-left of the body. It can be confirmed from the summation of the attention weight that the positional information is correctly captured (Figure 3E, lower-left).

In order to diagnose the normal condition from information on the chest radiograph, it is necessary to deny the possibility of any other abnormal findings. The abnormal part of each finding is distributed in various areas in the chest radiograph. Therefore, the entire image must be interpreted comprehensively. We confirmed the distribution of weights of the attention mechanism when the model generated a finding with ``\begin{CJK}{UTF8}{ipxm}異常なし\end{CJK} (English: normal)".  The attention mechanism was indicated to function over the entire region of the upper body in the image (Figure 3F).

The evaluation result of each test dataset is shown in Table 2. In the original distribution dataset, both oversampling and undersampling methods have shown a high BLEU score. In the dataset with only abnormal findings, undersampling scored slightly higher. Oversampling was able to generate a variety of findings compared to undersampling when there were many types of findings in the dataset (threshold is 5). However, undersampling was more variety when there were a few types of findings in the dataset (threshold is 30).

\begin{figure*}[ht!]
\begin{center}
\includegraphics[scale=0.45]{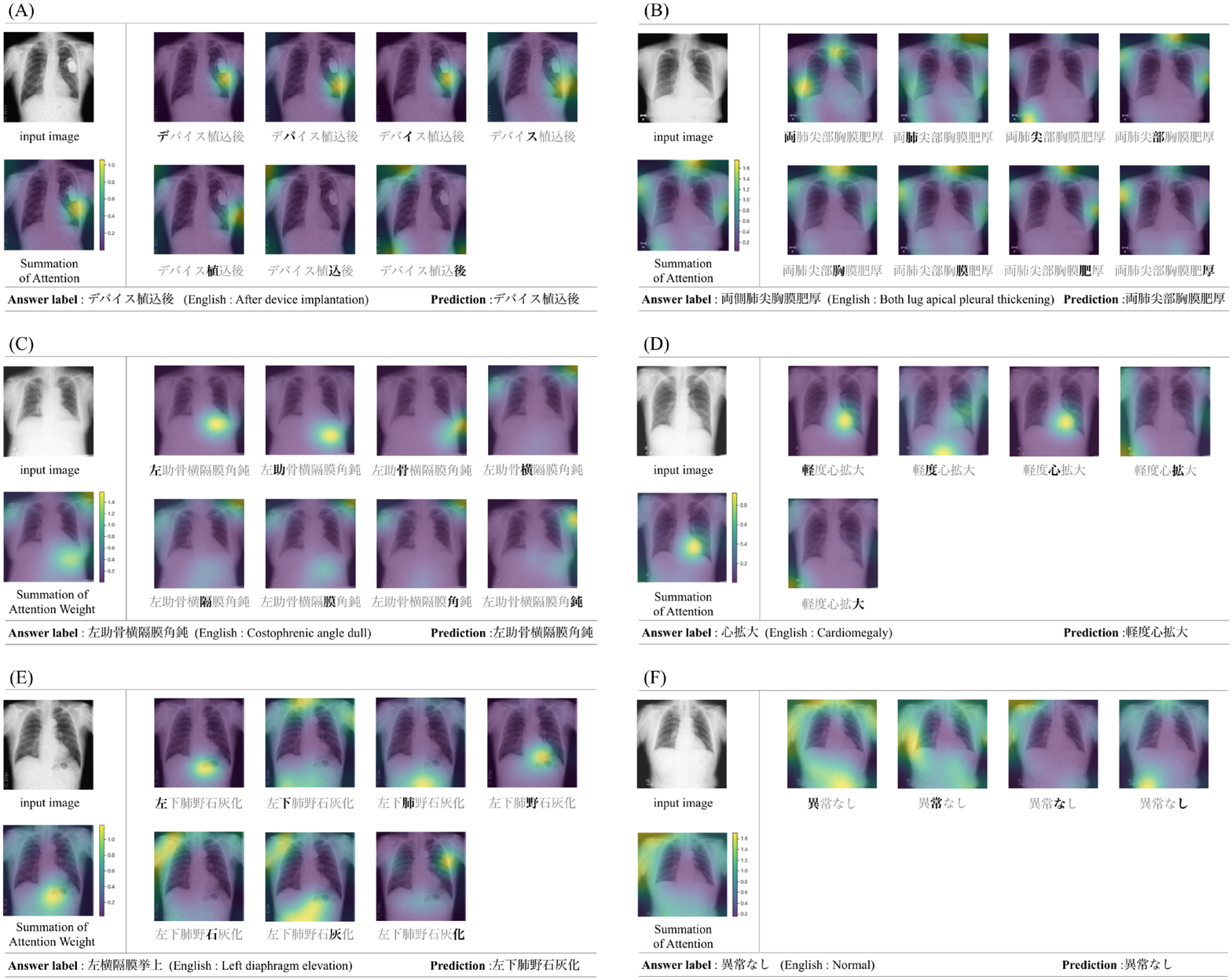}
\end{center}
\caption{Visualization of the attention mechanism and the generated medical findings. The upper-left image of each result represents the input image, and the lower-left image represents the summation of the weights of the attention mechanism generated at each step. The right side of each result shows the weight of the attention mechanism corresponding with the generated character. Each result below is the answer in the Japanese and English languages. In addition, the finding generated using the proposed method is also shown.}
\end{figure*}

\section{Discussion}
In the generation of the findings, it was confirmed that the attention mechanism was able to capture the existence of the device and positional information (Figure 3A, B, C). Positional information appears in the first half of the findings. The initial value of the decoder at the time of finding generation depends on the feature maps extracted from the encoder. The encoder needs to extract appropriate information. Therefore, pre-training using chest radiographs is important.  

There are orthographic variants in the findings of the medical images; thus, the existing evaluation metrics, such as the BLEU score are difficult to evaluate correctly. For example, ``\begin{CJK}{UTF8}{ipxm}両側肺尖胸膜肥厚\end{CJK}" and ``\begin{CJK}{UTF8}{ipxm}両肺尖部胸膜肥厚\end{CJK}" have the same meaning in terms of words (Figure 3B). However, the similarity of the two findings is low because the existing evaluation metrics consider the position of each character and do not consider the orthographic variants.

The proposed model achieved a high BLEU score in all the methods of the original distribution dataset (Table 1, top). The finding of ``\begin{CJK}{UTF8}{ipxm}異常なし\end{CJK} (English: normal)" accounts for 75.58\% of the total data in the original distribution dataset. Therefore, there is a possibility that the total BLEU score will be highly estimated by always generating ``\begin{CJK}{UTF8}{ipxm}異常なし\end{CJK} (English: normal)". For properly evaluating the abnormal findings, we evaluated the dataset with only abnormal findings. In the case of the dataset with only abnormal findings, the total BLEU score was lower. One of the reasons for this was the problem in the evaluation metrics that introduced example in the result section. The abnormal findings  are difficult to evaluate correctly with the BLEU score, because the findings have many orthographic variants.

We evaluated the accuracy of the data for ``\begin{CJK}{UTF8}{ipxm}異常なし\end{CJK} (English: normal)" with perfect match. The accuracies were 93.84\% and 76.99\% for the datasets with thresholds of 5 and 30, respectively. For the diagnosis of ``\begin{CJK}{UTF8}{ipxm}異常なし\end{CJK} (English: normal)", the model needs to deny the possibility of other abnormal findings. Therefore, the model is required to perform a comprehensive search of the entire image. Generating a finding of ``\begin{CJK}{UTF8}{ipxm}異常なし\end{CJK} (English: normal)" is difficult as a problem setting. From the results, the attention mechanism primarily functions on the abnormal part when the abnormal findings generated (Figure 3A, B, C, D, E). Conversely, the attention mechanism was widely functioning in the case of ``\begin{CJK}{UTF8}{ipxm}異常なし\end{CJK} (English: normal)" (Figure 3F).

We compared oversampling with undersampling. The model of oversampling was able to capture the characteristics of normal findings because similar number of normal findings and total abnormal findings were included in the dataset. However, the model of undersampling exhibited a lower BLEU score. The number of normal findings and each abnormal findings was the same in the dataset with undersampling. Problem setting is challenging with normal findings; therefore, training of the model requires more data than other findings. In the medical field, the model is required to consider abnormal findings. Therefore, it is generally important to increase recall over precision. From that perspective, undersampling was superior to oversampling in our experiments.

The proposed method can confirm the medical finding and the target region in the chest radiograph at the same time. Therefore, even if there is an error in the generated medical findings, it is to reduce the burden of checking on the medical expert by focusing on the region where the attention mechanism is functioning (Figure 3E). Currently, machine learning technology for medical images is often used for screening purposes to reduce the burden on medical health professional. Therefore, the interpretation ability for the output of the machine learning model is important.

\section{Conclusion}
In this paper, we proposed an end-to-end model that generates Japanese findings in character-level using an attention mechanism for the chest radiograph. Furthermore, the attention mechanism improved not only the accuracy, but also the interpretation ability of the results. In the evaluation, we used oversampling and undersamplling, common solutions for imbalanced data and discussed the characteristics of each method. As a feature application, the results of the attention mechanism can be used to help inexperienced medical specialists train interpret chest radiographs. Caption generation on a character-level has few practical examples because it is necessary to learn long-term relationships. However, character-level captioning is expected to be applied not only to medical images, but also to fields where there are numerous orthographic variants.  

Future work should define evaluation metrics with various field specialists that can be evaluated appropriately for the findings of medical images.

\section{Acknowledgments}
This research was supported by Japan Society for the Promotion of Science (JSPS) Grant-in-Aid for Scientific Research JP25700032, JP15H05327, JP16H06562 and Japan Agency for Medical Research and Development (AMED) of ICT infrastructure construction research business such as clinical research in 2016


\end{document}